\documentclass[10pt,twocolumn,letterpaper]{article}

\usepackage[pagenumbers]{wacv} 

\newcommand{\rev}[1]{{\color{black}#1}}

\definecolor{wacvblue}{rgb}{0.21,0.49,0.74}
\usepackage[pagebackref,breaklinks,colorlinks,allcolors=wacvblue]{hyperref}

\def\wacvPaperID{875} 
\def\confName{WACV}
\def\confYear{2026}

\usepackage{multirow}%
\usepackage[accsupp]{axessibility}
\title{GaussianHeadTalk: Wobble-Free 3D Talking Heads with Audio Driven \\Gaussian Splatting}

\author{
Madhav Agarwal$^{1}$ \quad
Mingtian Zhang$^{2}$ \quad
Laura Sevilla-Lara$^{1}$ \quad
Steven McDonagh$^{1}$ \\ 
$^{1}$University of Edinburgh \qquad
$^{2}$University College London \\
{\tt\small madhav.agarwal@ed.ac.uk, \ mingtian.zhang.17@ucl.ac.uk, \{l.sevilla,s.mcdonagh\}@ed.ac.uk}
}

\begin{document}
\twocolumn[{
\maketitle
\begin{center}
\vspace{-5pt}
\includegraphics[width=0.85\textwidth]{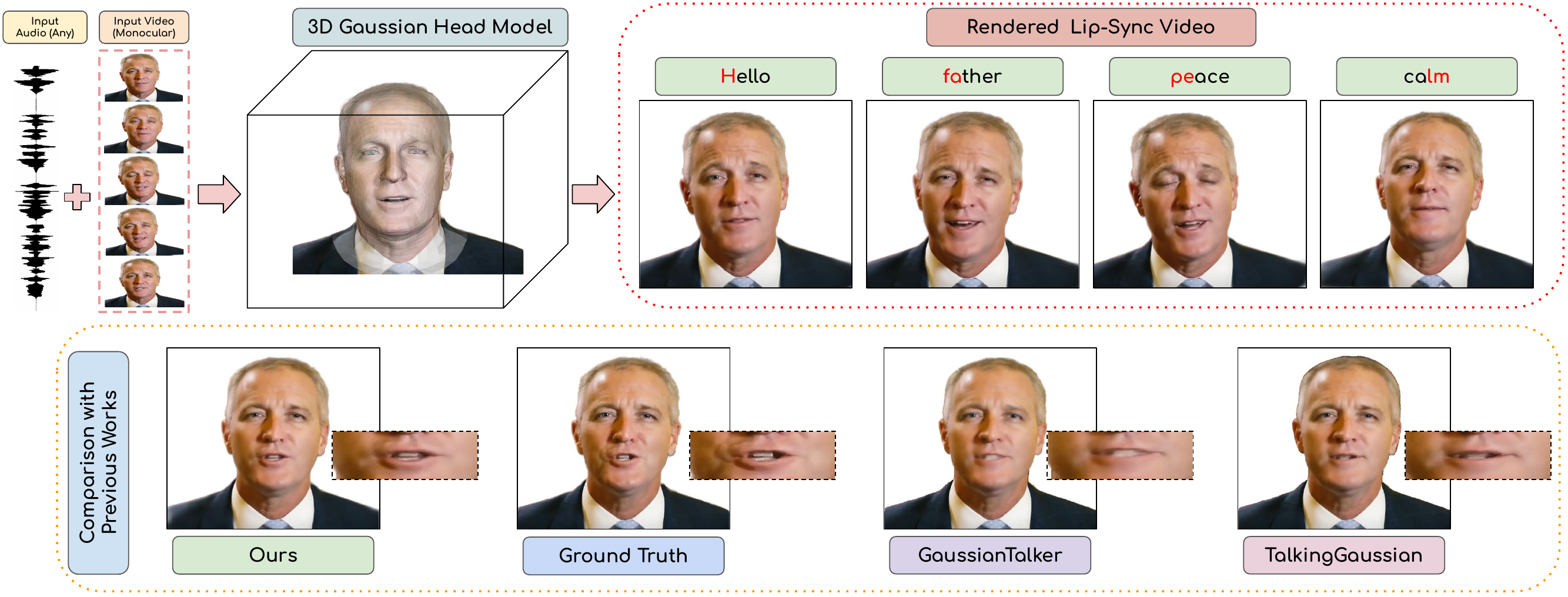}
\vspace{-5pt}
\captionof{figure}{To address the challenges of temporal instability, slow rendering, and limited photorealism in existing methods, we propose GaussianHeadTalk: a real-time system that generates photorealistic, temporally stable 3D talking head avatars directly from monocular video and arbitrary audio input.
Corresponding output frames generated by state-of-the-art methods GaussianTalker~\cite{cho2024gaussiantalker} and TalkingGaussian~\cite{li2024talkinggaussian} are also provided for visual comparison.}
\label{fig:teaser}

\end{center}
}]
\begin{abstract}
Speech-driven talking heads have recently emerged and  
enable interactive avatars. 
However, real-world applications are limited, as current methods 
achieve high visual fidelity
but slow or fast yet temporally unstable. 
Diffusion methods provide realistic image generation, yet struggle with one-shot settings. 
Gaussian Splatting 
approaches are real-time, yet inaccuracies in 
facial tracking, or inconsistent Gaussian mappings,
lead to unstable outputs and video artifacts that are 
detrimental to realistic use cases. 
%
We address this problem by mapping Gaussian Splatting using 3D Morphable Models to generate person-specific avatars. We introduce transformer-based prediction of model parameters, directly from audio, to drive temporal consistency. 
%
%
From monocular video and independent audio speech inputs, our method enables generation of real-time talking head videos 
%
%
where we report competitive quantitative and qualitative 
performance. 

\let\thefootnote\relax
\footnotetext{
    Project Page: 
    \href{https://madhav1ag.github.io/gaussianheadtalk}
         {https://madhav1ag.github.io/gaussianheadtalk}
}


\end{abstract}

\section{Introduction}
\label{sec:intro}
Generating talking head videos, driven directly by audio, can be considered a valuable task with multiple practical applications~\cite{compression2024review,compressing2022bmvc}. Whether in the education sector, health care, teleconferencing, or film and 
entertainment industries, high-quality personalized talking head avatars serve as an effective path for information transfer. For instance, AI-driven virtual assistants for telemedicine can be useful in assistive communications and post-stroke rehabilitation~\cite{afridi2025rehabilitation}. By providing a human face to 
an interactive agent, instead of a text-based input-output platform, the user experience is made more 
immersive~\cite{zhang2024unveilingimpactmultimodalinteractions}. Further uses include dubbing movies into multiple languages, which reduces the production time and cost of VFX studios manyfold~\cite{bigioi2023multilingual}. 

The canonical problem involves taking a short input video of a person, alongside an arbitrary speech audio signal, in order to create a person-specific avatar that can generate an output video of the subject appearing to speak the audio content (\emph{i.e.}\,with visual lip-syncing that accurately matches the input audio).
The task is commonly known as \emph{face reenactment} and previous solutions involve using GANs~\cite{Siarohin2019FOMM,wang2021facevid2vid,Agarwal2023AVFR,hong2022DAGAN}, Diffusion models~\cite{xu2024hallo,chen2024echomimic,wei2024aniportrait}, NeRFs~\cite{guo2021adnerf,Gafni_2021NeRFace} and, more recently, 3D Gaussian Splatting~\cite{cho2024gaussiantalker,li2024talkinggaussian,chu2024gagavatar,qian2024gaussianavatars}. 
Diffusion-based methods have 
robust generative priors and produce state-of-the-art image quality yet they are computationally expensive and inference speed is typically slower than GANs and NeRFs. 
In contrast, 3D Gaussian Splatting (3DGS) methods are person or scene-specific and have recently shown efficacy in rendering high-quality images and videos comparable to that of diffusion models, but at real-time speeds.
%

Although recent advancements in 3DGS have successfully incorporated temporal information for dynamic scenes~\cite{wu20244dGS, li2024spacetime}, the integration of related techniques into the synthesis of audio-driven 3DGS talking heads remains an open challenge. This gap highlights the need for novel approaches to combine dynamic, temporally consistent facial animation with audio-driven generation. 
%
The task is inherently dynamic, requiring precise temporal information to ensure realistic and consistent facial movements, particularly lip synchronization. Current audio-driven 3DGS methods rely on parameter tracking for temporal information, which often falls short for monocular videos. Inaccuracies in such tracking can lead to temporal flickering (i.e., `wobbling') in the face region, causing visible artifacts.
Our experiments show that this instability arises due to the improper utilization of
temporal information from an input video, which manifests itself as either inaccurate 3D mesh parameter tracking from RGB videos or independent frame-by-frame generation. 

To address this problem, we leverage a transformer architecture~\cite{vaswani2017attention} 
to process the audio signal in a manner that can capture long-range semantic information~\cite{peng2023selftalk,thambiraja2023imitator,song2024talkingstyle}. 
In tandem, we use the input video to learn a person-specific style embedding, which can maintain the visual identity of the speaker. We conjecture that directly mapping an audio signal to rasterised pixel space is highly challenging due to its high dimensionality, inherent non-linearity, and the extensive data required to cover the diverse output distribution of realistic facial appearances. We therefore alternatively opt to predict the FLAME~\cite{FLAME:SiggraphAsia2017} parameters for a template mesh and use them to render the subject head using 3DGS~\cite{qian2024gaussianavatars}. 
Although previous work has explored predicting 3DMM parameters from audio~\cite{faceformer2022,xing2023codetalker}, our novel architecture uniquely integrates a person-specific style embedding to preserve identity information, alongside direct FLAME parameter prediction from audio.
This direct prediction allows temporal information from the audio to inherently influence and constrain consecutive frame predictions,  significantly enhancing temporal consistency and reducing `wobbling'. We transfer the lip movement generated from our transformer model and head motion from the original video through an optimized set of FLAME parameters. 

One aspect that is widely assessed when judging the quality of generated videos is that of stability~\cite{roberto2022stablizationsurvey,guilluy2021videostabilization}. Intuitively: 
``\emph{the videos are stable}'' is a subjective statement. To formalize the notion, we propose a stability metric; towards quantifying video temporal stability (see Sec.~\ref{sec:methodology:quant_temp}).

\noindent Our contributions can be summarized as follows:
\begin{itemize}
    \item We highlight the utility of transformer-based prediction for person-specific 3D Morphable Model (3DMM) parameters, from input audio. Our approach enables a temporally consistent mesh-based subject rendering.
    \item We introduce a metric to quantify the temporal stability of synthetic talking head avatars. 
    \item Our overall pipeline, coined GaussianHeadTalk, achieves real-time video rendering, while maintaining competitive performance across both perceptual quality and video stability metrics.
    

\end{itemize}

\section{Related Work}
\label{sec:related_works}

\subsection{2D Talking Head Generation} 
Image generation and editing capabilities of 
modern generative models have inspired many practical applications including talking head synthesis. 
Early 2D-image based talking head methods ingest 
a single input image of a person and use GANs to drive video reenactment~\cite{Siarohin2019FOMM,wang2021facevid2vid,Agarwal2023AVFR,guo2024liveportrait,hong2022DAGAN,drobyshev2024emoportraits,tan2025edtalk}. 
These methods generally make use of an intermediate representation such as facial keypoints~\cite{Siarohin2019FOMM,wang2021facevid2vid,Agarwal2023AVFR,hong2022DAGAN,guo2024liveportrait} or latent vectors~\cite{tan2025edtalk,drobyshev2024emoportraits} to map motions to pixel space. 
3D Morphable Models (3DMM)~\cite{ji2021audio,thies2020neural,zhang2021hdtf} also provide an intermediate representation 
by mapping a 2D input image to 3D space and then back to 2D, 
affording control of head rotation. 
Imperfect mappings, however, can lead to a lack of identity preservation in resulting generated videos. 
Audio-driven methods ~\cite{prajwal2020wav2lip,zhou2020makelttalk,musetalk} focus on achieving accurate lip-sync, while head motion is generally non-deterministically learned from the training dataset. 

Given the superior generation quality of Diffusion methods~\cite{dhariwal2021diffusion}, in comparison to GANs, some researchers recently employed them for face reenactment~\cite{xu2024hallo,chen2024echomimic,wei2024aniportrait}. 
These methods provide better image quality, but inference is slow and computationally expensive, making them infeasible for real-time generation.
\rev{A recent line of works~\cite{li2024ditto, kim2025moditalker} introduce real-time generation, but the use of single input source images does not provide these models with temporal motion information. 
We conjecture that this leads to problems like unnatural head movement, stiff torso, and output quality is significantly dependent on identity features such as teeth and eye appearance within the single source frame.}
2D based methods also suffer from a lack of detailed 3D facial geometry information. This impedes external control over facial motion and consistency during head rotation.

\subsection{3D Talking Head Generation} With the advent of 3D rendering techniques such as NeRFs~\cite{mildenhall2021nerf} and Gaussian Splatting~\cite{kerbl20233dgs}, researchers have started to explore 
these 
methods to render talking heads. 
NeRF-based approaches~\cite{guo2021adnerf,Gafni_2021NeRFace} learn a radiance field from multiple input images of a single scene. 
The volumetric rendering is performed based on an input controlling signal \textit{e.g.,} audio. 
AD-NerF~\cite{guo2021adnerf} has an intertwined architecture that models the head and torso using two separate networks, limiting its flexibility.
The original NeRF architecture results in slow rendering speed (${<}1$ FPS on NVIDIA V100 GPUs~\cite{yan2024nerfspeed}), for talking head synthesis~\cite{shen2022learning,liu2022semantic}.  
Protrait4D~\cite{deng2024portrait4d} uses multi-view synthetic data to learn tri-plane representations and, subsequently, Protrait4D-v2~\cite{deng2024portrait4dv2} works on pseudo-multi-view videos. 
We note that these methods cannot perform real-time rendering.

Gaussian Splatting has emerged as an 
effective real-time rendering method 
via Gaussian optimization on input scene meshes. The input meshes are generated from 
monocular or multi-view videos.
GaussianAvatar~\cite{qian2024gaussianavatars} and GaussianHead~\cite{wang2023gaussianhead} use parametric models to control head motion. 
While the former binds Gaussians on a FLAME~\cite{FLAME:SiggraphAsia2017} mesh, such that every mesh triangle has at least one Gaussian, the latter uses a motion deformation field and tri-plane representation.
To enable motion, these renders can be conditioned directly on audio or driving video ~\cite{cho2024gaussiantalker,li2024talkinggaussian,chu2024gagavatar} to create talking heads. 
GaussianTalker~\cite{cho2024gaussiantalker} and TalkingGaussian~\cite{li2024talkinggaussian} both utilize a tri-plane representation and fuse an audio signal to predict the deformation offsets in an end-to-end approach.
\rev{GaussianSpeech~\cite{aneja2024gaussianspeech} and GaussianTalker~\cite{yu2024gaussiantalker} use  FLAME~\cite{FLAME:SiggraphAsia2017} as an intermediate representation to map audio to Gaussians. GaussianSpeech~\cite{aneja2024gaussianspeech} focuses on generating high-dimensional vertex offsets from audio for multiview videos. 
GaussianTalker~\cite{yu2024gaussiantalker} predicts fine-grained offsets for Gaussian position, rotation, and color to synthesize details like teeth and wrinkles,  
however layering this detail-synthesis network on top of fully audio-generated motion may risk amplifying any underlying instability from the motion prediction module. 
These methods are suitable for real-time inference due to high rendering speed; however, we conjecture that independent frame-by-frame generation and a lack of optimization, using objectives that account for temporal tracking, have the potential to induce jittering artifacts.}
Another line of work directly predicts 3D Morphable Model (3DMM) parameters, such as FLAME~\cite{FLAME:SiggraphAsia2017}, from an audio signal~\cite{VOCA2019,richard2021meshtalk,faceformer2022,xing2023codetalker}. Their focus is on controlling facial parameters, rather than handling texture information, and hence, 
provide semantically meaningful motion controls. 
In this work, we take advantage of an intermediate 3DMM representation by mapping audio-to-face parameters and then render a video using Gaussian Splats with real-time performance.

\begin{figure*}[t!]
  \centering
  \includegraphics[width=\linewidth]{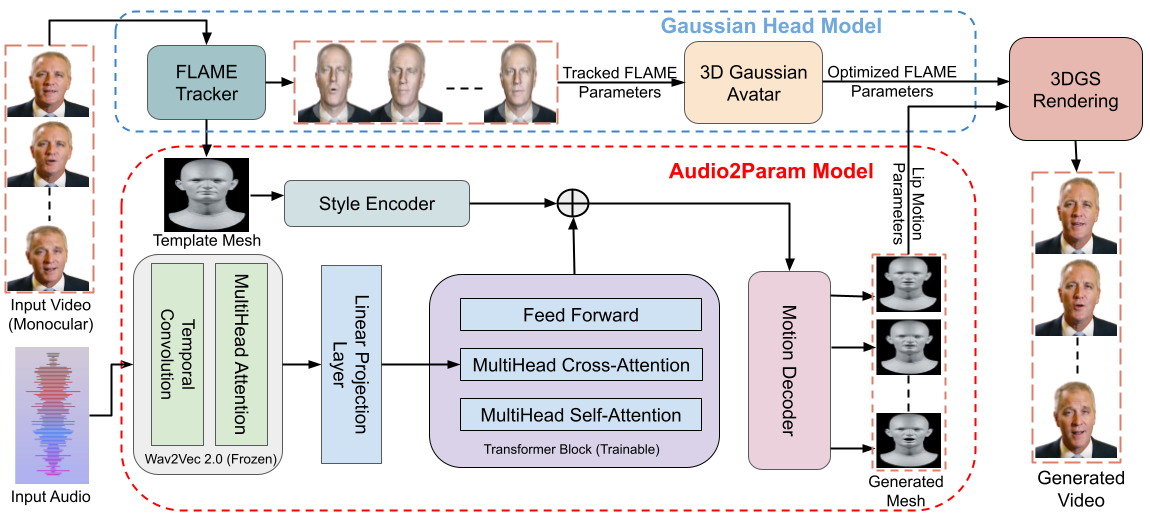}
  \caption{We introduce GaussianHeadTalk, which comprises of Gaussian Head Modeling and audio to facial motion mapping. We first generate meshes from an input video using VHAP~\cite{qian2024vhap} tracking. Given an input audio and a template mesh, The audio to facial motion mapping uses a transformer-based architecture with a frozen Wav2Vec 2.0~\cite{baevski2020wav2vec2} encoder. It learns long-term audio context and maps it directly to the 3D mesh by predicting FLAME~\cite{FLAME:SiggraphAsia2017} parameters. The generated parameters are used to render a person-specific GaussianAvatar~\cite{qian2024gaussianavatars}, trained using the input video.}
  \label{fig:architecture}
\end{figure*}

\section{Methodology}
\label{sec:methodology}


Our method is trained using an identity-specific video $V=\{I_n\}$, consisting of $n$ image frames. 
We build our model in two-stages, where the first stage involves training identity-specific Gaussian Splatting from the input video $V$, such that each Gaussian is optimized with respect to a 3D Morphable Model's triangles by ensuring that every triangle is attributed to at least one Gaussian. 
The first stage of our pipeline (see Sec.~\ref{sec:methodology:GaussianHeadModel}) builds upon GaussianAvatar~\cite{qian2024gaussianavatars}, where we replace the original FLAME~\cite{FLAME:SiggraphAsia2017} parameters with parameters optimized by person-specific avatar training. 
In the second stage (Sec.~\ref{sec:methodology:Audio2Param Model}), we learn an audio to FLAME~\cite{FLAME:SiggraphAsia2017} 
mapping, which captures the speech style of a given identity. We next provide details for each stage.

\subsection{Gaussian Head Model}
\label{sec:methodology:GaussianHeadModel}

3D Gaussian Splatting (3DGS)~\cite{kerbl20233dgs} reconstructs a static scene in 3D space using images and intrinsic, extrinsic camera parameters. 
A scene is represented using a set of $K$ anisotropic 3D Gaussians, where each Gaussian is defined by a center mean $\mu_i \in \mathbb{R}^3$ and a covariance matrix $\Sigma_i \in \mathbb{R}^{3 \times 3}$. 
The density of the $i$-th Gaussian for a 3D coordinate $x\in \mathbb{R}^3$ is given by:
\begin{equation}
\label{eq:gaussian}
   G_i(x) = e^{-\frac{1}{2}(x - \mu_i)^\top \Sigma_i^{-1} (x - \mu_i)}.
\end{equation}
\\Further decomposing the covariance matrix for efficient storage and rendering, we obtain \mbox{$\Sigma = R S S^\top R^\top$} where $R$ is a rotation matrix and $S$ a scaling matrix. By additionally storing appearance information, a 3D scene can be defined by a set of 3D Gaussian primitives:

\begin{equation}
\label{eq:3Dscene}
\mathcal{G} = \left\{ G_i = (\mu_i, s_i, q_i, \alpha_i, \text{SH}_i) \right\}_{i=1}^K,
\end{equation}
where $\mu_i \in \mathbb{R}^3$ is the position vector (c.f.~Eq.~\ref{eq:gaussian}), $s_i \in \mathbb{R}^3$ is the scaling vector, $q_i \in \mathbb{R}^4$ is a quaternion representing orientation, $\alpha_i \in \mathbb{R}$ is an opacity value, and $\text{SH}_i$ denotes a set of spherical harmonics for encoding color as a function of view direction.

At rendering time, the 2D pixel-wise color ${C}$ is calculated by blending a subset of all 3D Gaussians whose projection into the image plane overlaps with that pixel location.
Let $\mathcal{N} \subseteq \{1, \dots, K\}$ denote the set of overlapping Gaussians:

\begin{equation}
    \text{C} = \sum_{i\in \mathcal{N}} c_i \alpha_i' \prod_{j=1}^{i-1} (1 - \alpha_j'),
    \label{eq:1}
\end{equation}
\\where $c_i$ is the view-dependent color of the $i$-th Gaussian, and $\alpha_i'$ is the projected 2D opacity of each Gaussian, obtained by multiplying the projection of the overlapping 3D Gaussian onto the image plane with the original opacity $\alpha$.

GaussianAvatar~\cite{qian2024gaussianavatars} introduce a method to bind Gaussians to Morphable Model mesh triangles, in this case, a FLAME~\cite{FLAME:SiggraphAsia2017} representation. 
For a given triangle with vertices and edges, a Gaussian is initialized using the mean position of the vertices, the direction of one edge, and the normal vector of the triangle. 
A process of Gaussian densification helps to adjust to an appropriate number of Gaussians, based on local scene complexity. This involves increasing or decreasing the number of Gaussians in a given part of the scene and is achieved by either splitting Gaussians into two if the view-space positional gradient is large, or cloning into two if it is small. 
To avoid density explosion, a pruning strategy removes points that have very low opacity, while maintaining at least one splat per triangle.

The stability of the rendering process depends heavily on the accuracy of the binding between Gaussian splats and FLAME~\cite{FLAME:SiggraphAsia2017} triangles. 
In contrast to alternative work, such as INSTA~\cite{zielonka2023insta}, where bounding volume hierarchy (BVH) based nearest triangle search~\cite{clark1976bvh} 
leads to flickering artifacts, GaussianAvatar~\cite{qian2024gaussianavatars} is agnostic to tracked mesh inaccuracies due to back-propagation of a positional gradient for each triangle.
This consistent binding between Gaussians and the mesh triangles, regardless of pose or expression, allows fine-tuning of FLAME parameters.
Along with the optimization of Gaussian splats parameters for position and scaling, FLAME parameters (translation, pose, and expression) were therefore also optimized during training. 
This plays a crucial role in stabilizing the rendering output, mitigating misalignment between the triangle meshes and Gaussians. We leverage this Gaussian-based head modeling and FLAME parameter tuning to help generate stable output.


\begin{table*}[ht]
    \centering
    \begin{tabular}{l|ccccc|cc|c}
    \hline
    \multirow{2}{*}{Method} & \multicolumn{5}{c|}{Self-Reenactment} & \multicolumn{2}{c|}{Cross-Reenactment} & \multirow{2}{*}{FPS$\uparrow$}\\ \cline{2-8} 
     & PSNR$\uparrow$ & SSIM$\uparrow$ & LPIPS$\downarrow$ &  Sync$\uparrow$ & Stability$\downarrow$ & Sync$\uparrow$ & Stability$\downarrow$ \\ \hline
    IPLap~\cite{Zhong2023iplap} & 29.0412 & 0.9462 & 0.0340 &  3.902 & 0.6633 & 3.324 & 0.6856 & 3.4\\ 
    EDTalk~\cite{tan2025edtalk} & 26.9461  & 0.8626  & 0.0486 & \textbf{7.144} & 0.7802 & \textbf{6.982}& 0.7931 & 17.2\\ \hline
    \rev{Ditto~\cite{li2024ditto}} & 21.0595 & 0.7412  & 0.1284  & 7.023  & 0.9245 & 6.844 & 0.9618 & 24.2 \\ \hline
    MimicTalk~\cite{ye2024mimictalk} & 23.8775 &  0.8092 & 0.0735  & 5.446  & 0.8824 & 5.286 & 0.9227 & 12.1\\ \hline
    GaussianTalker~\cite{cho2024gaussiantalker} & 27.6079 & 0.9352  & 0.0451  & 5.346 & 1.7622 & 5.042 & 1.8745 & 59.6\\ 
    TalkingGaussian~\cite{li2024talkinggaussian} & 27.3053  & 0.9335 & 0.0342 & 6.422 & 1.7183 & 6.146 & 1.8803 & \textbf{72.2}\\ \hline
    \textbf{GaussianHeadTalk} & \textbf{29.1233}  & \textbf{0.9477} & \textbf{0.0338} & 6.528 & \textbf{0.6201} & 6.122 & \textbf{0.6836}  & 45.4 \\ \hline
    \end{tabular}
    \caption{Self-Reenactment and Cross-Reenactment experimental settings. Our method achieves strong results in terms of stability, realism, image quality and remains competitive for lip-sync.}
    \label{tab:quantitative_results}
\end{table*}

\subsection{Audio to Facial Motion (Audio2Param)} 
\label{sec:methodology:Audio2Param Model}
We map from an audio signal to facial motion by leveraging the FLAME parametric 3D Morphable Model. FLAME disentangled parameters control identity, expression, and pose. These parameters can then be used to generate an explicit 3D head mesh.
Distinct from previous work~\cite{faceformer2022,xing2023codetalker}, which operates directly on full 3D head meshes by predicting triangle deformations or vertex positions, we take advantage of the disentangled FLAME representation. 
By directly predicting FLAME expression parameters, 
we reduce the complexity of our learning objective from explicitly predicting the spatial location of 
thousands of face vertices to the prediction of fewer than one hundred parameters that together define facial expressions and lip motion. 

We design a transformer-based architecture to capture long-range temporal information from the audio signal concerning the context of the spoken sentence. 
To mitigate the lack of diverse 3D audio-video datasets containing explicit visual data, 3D meshes and paired audio, we instantiate our encoder using Wav2Vec 2.0~\cite{baevski2020wav2vec2} which has previously demonstrated strong representation performance for audio information~\cite{faceformer2022,xing2023codetalker}. 
%
%
We encode audio signals into feature vectors by adding a linear projection layer after the encoder. 
Similar to~\cite{faceformer2022}, we use a Periodic Positional Encoding (PPE) to provide temporal information to the transformer decoder and a binary alignment mask to avoid information leakage from future frames.

For a single identity $m$, let the input training set be given by $L = \{A,M_{{gt}}^{0:T},N_m\}$, where 
$M_{gt}^{0:T}$ is a sequence of ground-truth meshes for $T{+}1$ frames, $A$ is an audio signal from the ground-truth video that corresponds to those frames. 
The neutral template mesh $N_m$ represents the given identity. Each input training set is generated by processing an input video consisting of $T{+}1$ frames using the VHAP tracker~\cite{qian2024gaussianavatars} to generate ground truth meshes $M^{0:T}_{gt}$ and neutral template mesh $N_m$.
Our objective is to predict a sequence of meshes $M^{0:T}_{pred}$, given audio and neutral template mesh, such that:
\begin{equation}
    f_{\theta}(A,N_m) = M^{0:T}_{pred} \approx M^{0:T}_{gt}
    \label{eq:objective}.
\end{equation}

The correlation between the audio signal and lip movement is typically high, but the correlation between the audio signal and head movement is not~\cite{yehia1998quantitative}. Since different yet plausible head motions and expressions exist for the same speech, there is no one-to-one mapping between speech and head motion. 
Hence we focus on predicting accurate lip movement from audio, by checking for high correlation between audio and lip movement, 
and transfer head motion directly from the original tracked video sequence. 

In addition to the head motion, the (potentially independent) audio speech signal $A$ is processed through the transformer encoder and linear projection layer. We denote the output from the linear projection layer for $T{+}1$ frames as  $C^{0:T}$. For a given frame $t$, the transformer encoder takes audio for frames $[0,\ldots,t]$ and uses the linear projection layer to generate $C^{t}$.
The predicted audio features are passed to the multi-head attention block of the transformer to obtain the latent vertex offsets $O_v^{0:T}$ for each frame.  

To utilize latent vertex offsets, an identity-specific template mesh, which is an average of all meshes obtained through video tracking, is encoded through a style encoder network to obtain an identity embedding $S$. This procedure is shown in Fig.\,\ref{fig:architecture}. Predicted latent vertex offsets $O_v^i$ for frame $i$ are linearly combined with this embedding as: 
\begin{equation}
    O_{sv}^i = S+ O_v^i, \quad i  \in \{0, \ldots, T\}
    \label{eq:mesh_predicted}.
\end{equation}

The style-conditioned latent embeddings  $O_{sv}^{0:T}$ are then processed by a motion decoder, which comprises a set of linear layers that map them to a low-dimensional FLAME parameter space, to obtain a 3D mesh representation. By performing this process for each frame $i$, we obtain a predicted mesh sequence $M^{0:T}_{pred}$.     

Toward achieving accurate lip motion and jaw movement prediction, we isolate the FLAME parameters responsible for jaw movement. We use these to augment the ground-truth mesh as $M_{gt'}$ 
and calculate a loss as the difference, in vertex space, between this and our predicted mesh per frame. The remaining FLAME parameter values, used to define the ground-truth mesh, are instantiated using the template mesh. 
%
The model is trained end-to-end using an $L2$ loss between the ground-truth and predicted meshes in vertex space as follows 
\begin{equation}
\mathcal{L}_{mesh} = \sum_{n=1}^{N} \left( \sum_{t=1}^{T} \left\| M_{gt'}^t - M_{pred}^t \right\|_2 \right)
\label{eq:vertex_loss}.
\end{equation}

\noindent
During inference, the Audio2Param component ingests a neutral mesh and audio signal in order to predict a sequence of animated 3D facial meshes using the FLAME parameter space. The predicted FLAME parameters are used to drive the motion of a person-specific avatar~\cite{qian2024gaussianavatars}, 
culminating in the generation of an audio driven talking head. 




\subsection{Quantifying temporal consistency}
\label{sec:methodology:quant_temp}
Noted existing work generate talking head videos by posing video rendering as a set of, per-frame, independent tasks. 
We observe that this typically leads to a lack of temporal consistency in the output, which manifests itself as unnatural wobbling, aberrations, and facial oscillations. Toward quantifying this problem, we employ a
measure of temporal smoothness for a given video.

\begin{figure}[htbp]
  \centering
  \includegraphics[width=\linewidth]{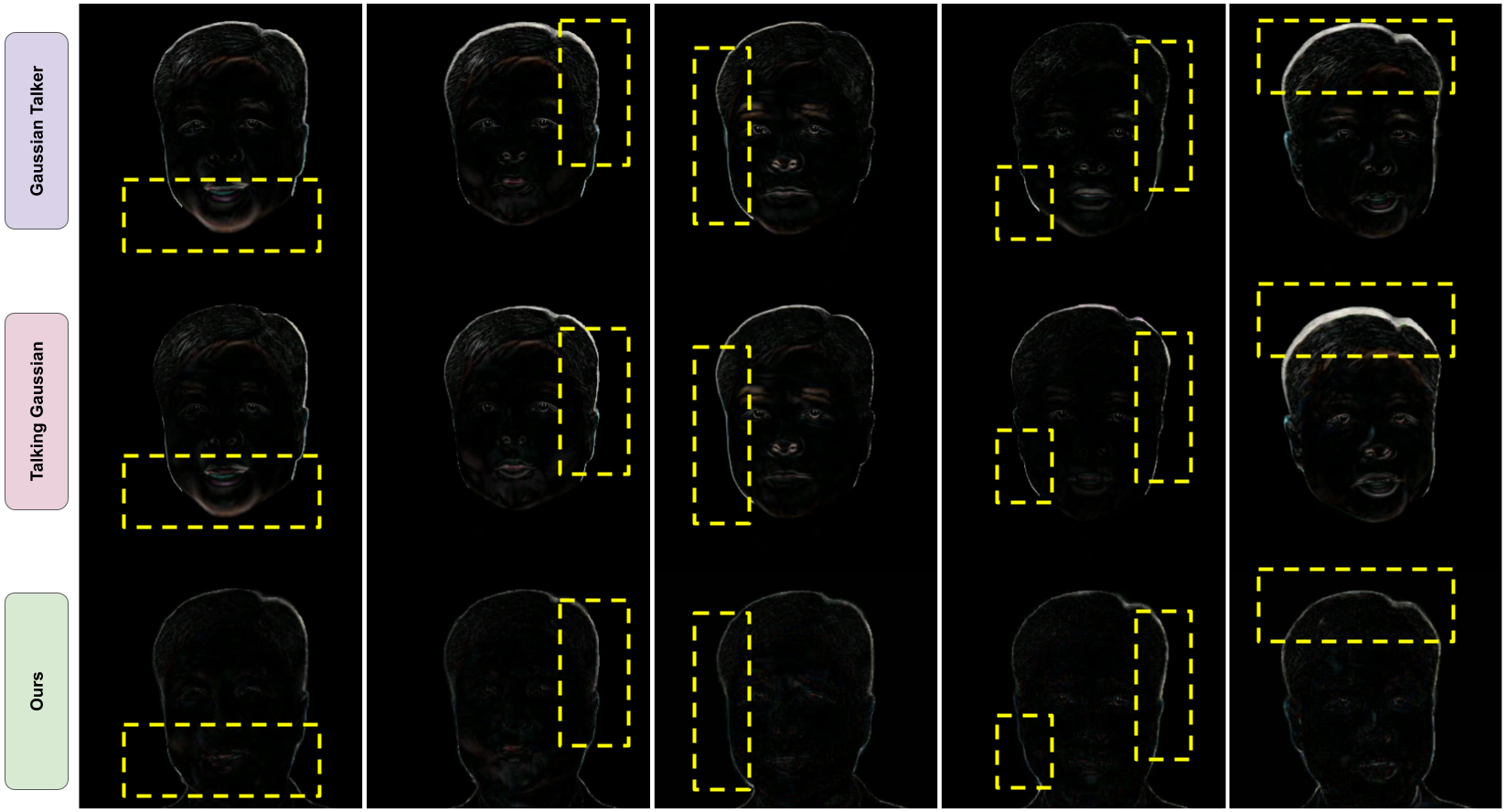}
  \caption{Facial ``wobbling'' artifacts can be visualized via the absolute difference between generated and ground-truth frames, in image-space. The temporal gap between consecutive columns is ten frames in each case. Our method exhibits smaller and spatially-more-stable disparities, across time.}

  \label{fig:wobbling-overlay}
\end{figure}

\begin{figure}[t!]
  \centering
  \includegraphics[width=\linewidth]{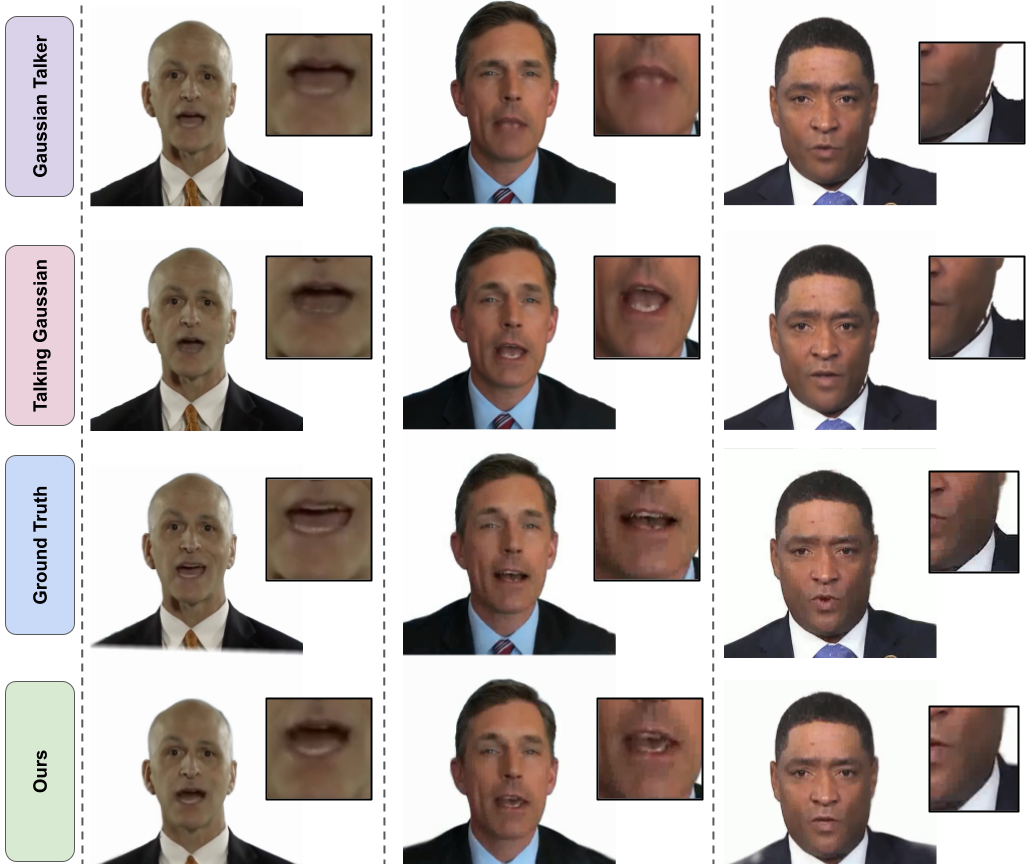}
  \caption{Self-Reenactment Results: We show qualitative results between the Gaussian based methods by reenacting them using  the original audio. Our method, GaussianHeadTalk, shows better mouth movement, sharper teeth and fewer artifacts.}
  \label{fig:self-reenactment}
  \vspace{-6pt}
\end{figure}

We first select a video and accompanying audio sample from the dataset~\cite{zhang2021hdtf} and proceed to render a talking head video using the original audio signal. This enables a direct comparison between the generated video and the original video (ground-truth). Towards defining a robust evaluation protocol, we detect and track facial key points~\cite{Zhou2023STAR} on the nose, as these key points are largely unaffected by jaw movement and expression changes. The time-domain signal provided, by these key points, can then be compared between generated and ground truth video frames. 
Further, we observe that high-frequency wobbling and rapid oscillations are challenging to detect using keypoint comparisons alone, and thus adopt a hybrid approach by additionally performing a Fast-Fourier-Transform (FFT) analysis to identify frequent and uneven oscillations. 


Our hybrid approach takes an average of mean motion difference $M_d$, variability in motion magnitude $V_m$, and high-frequency power $H_f$. Each term is normalized by their respective maximum values across a given sequence of input frames. 
Our compound stability score is then calculated by taking the average of these values:
\begin{equation}
    \text{Stability score} = \frac{M_{d} + V_{m} +  H_{f}}{3}
    \label{eq:weighted_sum}.
\end{equation}

\section{Experiments}
\label{sec:experiments}

\begin{figure*}[t!]
  \centering
  \includegraphics[width=\linewidth]{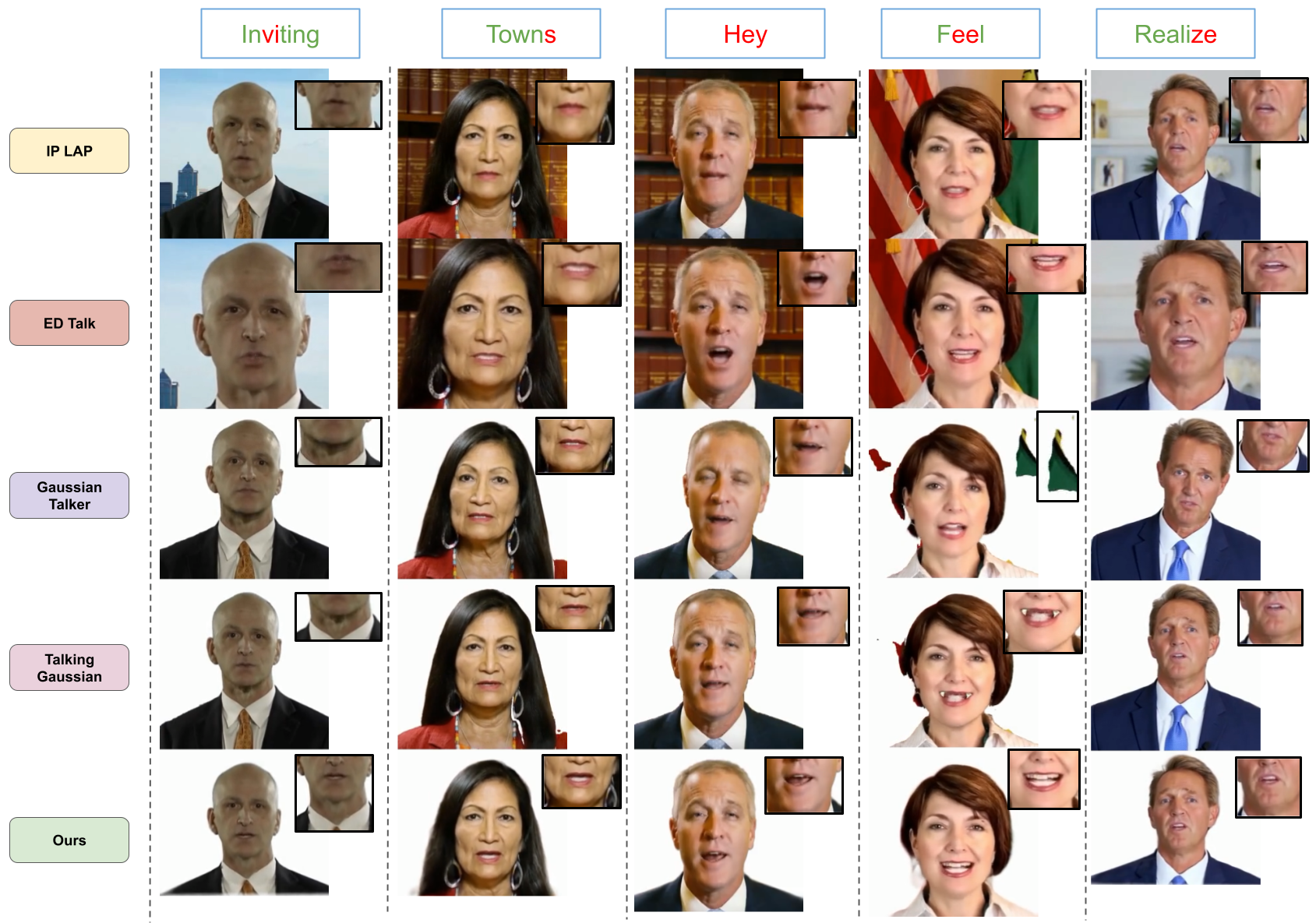}
  \caption{Cross-Reenactment Results: visual reenactment using various methods with audio from a different speaker. The top row shows the words from the audio, where red text highlights exact phonemes. GaussianHeadTalk can provide 
  improved lip movement for these audio samples where other methods struggle 
  with lip motion.
  }
  \label{fig:cross-reenactment}
  \vspace{-12pt}
\end{figure*}

\subsection{Experimental Settings}
\label{sec:experiments:settings}
\paragraph{Dataset:} 
We perform experiments on two datasets: VOCASET~\cite{VOCA2019} and HDTF~\cite{zhang2021hdtf}. Both datasets provide videos and synchronized audio. VOCASET also provides tracked 3D-scans of the faces. We use VOCASET for pre-training the Audio2Param component, and HDTF for training and evaluating person-specific gaussian avatars. Since the Gaussian Splatting and NeRF-based models require subject specific training, we select ten subjects from HDTF that cover a diverse set of identities and have a minimum of four minutes in video length. All videos were converted to 25fps to maintain experimental consistency. 
We synthetically generate 15 audio clips covering five different languages, with an average duration of seven seconds, using a text-to-speech model\footnote{https://elevenlabs.io/}.

\paragraph{Comparison Baselines:} 
We compare GaussianHeadTalk with current state-of-the-art methods. 
Two approaches, GaussianTalker~\cite{cho2024gaussiantalker} and TalkingGaussian~\cite{li2024talkinggaussian}, naturally align with our proposed problem setting as both can be considered audio-driven Gaussian methods. We also compare with IP LAP~\cite{Zhong2023iplap} and ED Talk~\cite{tan2025edtalk} which are GAN based methods, \rev{and Ditto~\cite{li2024ditto} which is a Diffusion based method}. Lastly, we use MimicTalk~\cite{ye2024mimictalk} to evaluate performance of a related NeRF technique.

\paragraph{Implementation Details:} Our method is built on PyTorch. We first used VOCASET~\cite{VOCA2019} audio-video and their  tracked FLAME~\cite{FLAME:SiggraphAsia2017} parameters. 
We train our Audio2Param component using ADAM~\cite{kingma2014adam} optimizer with a learning rate of 1$e^{-4}$. We pre-train for 
50000 steps and all experiments were performed on a single NVIDIA Tesla A100 GPU (40GB).

\subsection{Quantitative Evaluation}
\label{sec:experiments:quant}
We evaluate the performance of our model on two tasks: self-reenactment and cross-reenactment. First, for self-reenactment, we extract the first 30 seconds of a video as a test set. We train on the remaining part of the video segment. For cross-reenactment, we use synthetically generated audio from a text-to-speech model so that the audio sample contains no information about the trained person identity. We compare our method with state-of-the-art Gaussian Splatting~\cite{cho2024gaussiantalker,li2024talkinggaussian}, GAN ~\cite{Zhong2023iplap,tan2025edtalk}, \rev{Diffusion~\cite{li2024ditto}} and NeRF~\cite{ye2024mimictalk} based methods. To evaluate self-reenactment, we use Peak Signal-to-Noise Ratio (PSNR), Structural
Similarity Index Measure (SSIM), and Learned Perceptual Image Patch Similarity (LPIPS). We calculate the Sync confidence score~\cite{chung2017sync,chung2017syncnet} for both self-reenactment and cross-reenactment. We observe that our method predominantly improves upon the state-of-the-art (ref.~Table~\ref{tab:quantitative_results}). For the perceptual metric GaussianHeadTalk performs better than NeRF and alternative Gaussian based methods. IPLap~\cite{Zhong2023iplap} provides results comparable to ours, as it models only the lip region using a GAN-based architecture. However, inference is somewhat slower, which may impede its real-time applications 
(ref.~Table~\ref{tab:quantitative_results}). 


\subsection{Qualitative Evaluation}
\label{sec:experiments:qual}
We show visual results in Figure~\ref{fig:self-reenactment} and Figure~\ref{fig:cross-reenactment} for qualitative comparison. GAN based methods provide good lip-sync in both cases, but their image quality falls short; with generated videos of resolution up to $256\times256$. Gaussian-based methods (GaussianTalker, TalkingGaussian) generate sharper images, but their lip sync scores are low. As example, they show lower lip openness while speaking `Hey', (see middle column of Fig.~\ref{fig:cross-reenactment}).  They also display wobbling artifacts in the generated videos, mainly due to the lack of long-term temporal information and improper tracking of 3D parameters during training. Our method generates stable talking head videos, with qualitative results that concur with the relative quantitative metric improvements. We provide supplementary videos for further results visualization.

\subsection{User Study}

We conduct a user study to investigate how generated video quality is perceived by humans. 
We select a group of \rev{thirty} individuals and present each survey participant with multiple video triplets. 
Our study compares the performance of GaussianTalker~\cite{cho2024gaussiantalker}, TalkingGausian~\cite{li2024talkinggaussian} and our work. 
Participants were asked to evaluate videos in terms of ``naturalness'' (\ie assess wobbling and artifacts), lip sync quality, and image quality (\ie evaluate identity preservation in generated videos). 
Generated video orderings were randomized and models responsible for generation were masked from participants. 
Each participant was shown ten sets of video triplets, each with an average duration of five seconds.
Participants were asked to provide an ordinal ranking for each triplet, for each assessed aspect: `Best', `Average', and `Worst', which we numerically map to values 3, 2, and 1, respectively. 
\rev{For each participant, we sum the ratings a method received across the 10 triplets, and then divided this sum by 3 to normalize the score into a range of 3.3–10. 
The final reported scores, for each method, are average normalized scores across all thirty participants (Table~\ref{tab:user_study}). }

\label{sec:experiments:user}
\begin{table}[htbp]
    \centering
    \resizebox{\linewidth}{!}{%
    \begin{tabular}{|l|c|c|c|}
    \hline
    Method & Natural$\uparrow$ & LipSync$\uparrow$ & Quality$\uparrow$\\ \hline
    GaussianTalker~\cite{cho2024gaussiantalker} & 4.0 & 5.0 & 4.0\\ \hline
    TalkingGaussian~\cite{li2024talkinggaussian} & 6.2 & 7.2 &  6.5\\ \hline
    GaussianHeadTalk (ours) & \textbf{9.8} & \textbf{7.8} & \textbf{9.5} \\ \hline
    \end{tabular}
   } 
    \caption{User Study assessing human visual perception of generated video quality. The scores are averaged over different participants, with ten being the maximum.}
    \label{tab:user_study}
\end{table}

All participants ranked videos generated by our method as the most natural, which supports our improved video stability claims. 
In terms of `LipSync', our method scores slightly above TalkingGaussian, which correlates with the related `Sync' score (Table~\ref{tab:quantitative_results}). Performant lip sync quality may be explained by the special focus on the lip region, in both cases. 
Image quality for our method can also show improved human rating scores, with respect to the compared state-of-the-art. 


\subsection{Ablation Study}
\label{sec:experiments:ablate}
We perform ablative studies to evaluate our methodological choices (see Table~\ref{tab:ablation}). We first explore the effect of using non-person-specific (\ie non-optimized) FLAME parameters directly for rendering (``w/o Parameter Optimization''). Resulting generated videos display artifacts around various parts of the face 
which arise due to inaccuracies in parameter alignment, distorting the videos generated. 

We also tested the effect of more restrictive motion transfer; namely, transferring only the lip motion (FLAME model jaw parameters) and keeping the head motion static (``w/o Full Motion Transfer''). 
This strategy leads to artifacts around parts of the generated video, due to the movement interdependence between distinct FLAME parameters. 
We observe smoother video generation, with fewer artifacts, when we transfer the full set of FLAME parameters (remaining pose and expression components) from the original video. 


\begin{table}[htbp]
    \centering
    \resizebox{\linewidth}{!}{%
    \begin{tabular}{|l|c|c|c|}
    \hline
    Method & PSNR$\uparrow$ & Sync$\uparrow$ & Stability$\downarrow$\\ \hline
    w/o Parameter Optimization & 27.6987 & 6.5123 & 1.1432 \\ \hline
    w/o Full Motion Transfer & 23.5621 & 6.4962 & 0.9154\\ \hline
    GaussianHeadTalk (ours) & \textbf{29.1233} & \textbf{6.528} & \textbf{0.6201} \\ \hline
    \end{tabular}
   } 
    \caption{Ablation study: removal of key method components results in qualitative visual degradations and respective decreases in associated metrics.
    }
    \label{tab:ablation}
\end{table}


\section{Limitations and Discussion}
Our method can generate high-quality talking heads, but is restricted to exactly this part of the body and currently cannot render partial or full human bodies. 
This limitation arises due to our usage of a head-specific parametric model. Extension to accommodate full-body reenactment might involve designing a Gaussian Splatting model that binds to SMPL-X \cite{SMPL-X:2019}, or similar full-body 3D parametric models.
We conjecture that a further interesting line of future work will involve exploring any benefits derivable from learning facial expression changes based on the \emph{tone} and \emph{speed} of the audio signal, or other external control parameters for changing the emotions of the face. 

\section{Ethical Consideration} 
\label{sec:ethical}
The generation of photo-realistic talking heads is a technology that carries potential risks of misuse, particularly in the creation of deepfakes for misinformation, harassment, or identity fraud. 
We advocate for the incorporation of robust watermarking and detection mechanisms to help distinguish synthetic content from real media and reduce the potential for harmful misuse.

\section{Conclusion}
\label{sec:conclusion}
We introduce a novel method for generating high-quality 3D talking heads with lip sync in real time. We proposed a temporally stable pipeline that uses transformers to capture sematic information and long-range dependencies from audio signals. 
We also introduce a stability metric to quantify perceptual wobbling in generated videos. Our method offers strong performance with respect to the existing state-of-the-art in terms of qualitative and quantitative benchmarks and we believe that high-quality real-time facial reenactment holds exciting potential for many practical and useful real-time applications.


\clearpage
{
    \small
    \bibliographystyle{ieeenat_fullname}
    \bibliography{biblio}
}
\clearpage
\setcounter{page}{1}
\maketitlesupplementary
\renewcommand{\thesection}{\Alph{section}}
\setcounter{section}{0}
\renewcommand{\thefigure}{\Alph{figure}}
\setcounter{figure}{0}
\renewcommand{\thetable}{\Alph{table}}
\setcounter{table}{0}
 
\begin{figure}[t!]
  \centering
  \includegraphics[width=\linewidth,height=300pt]{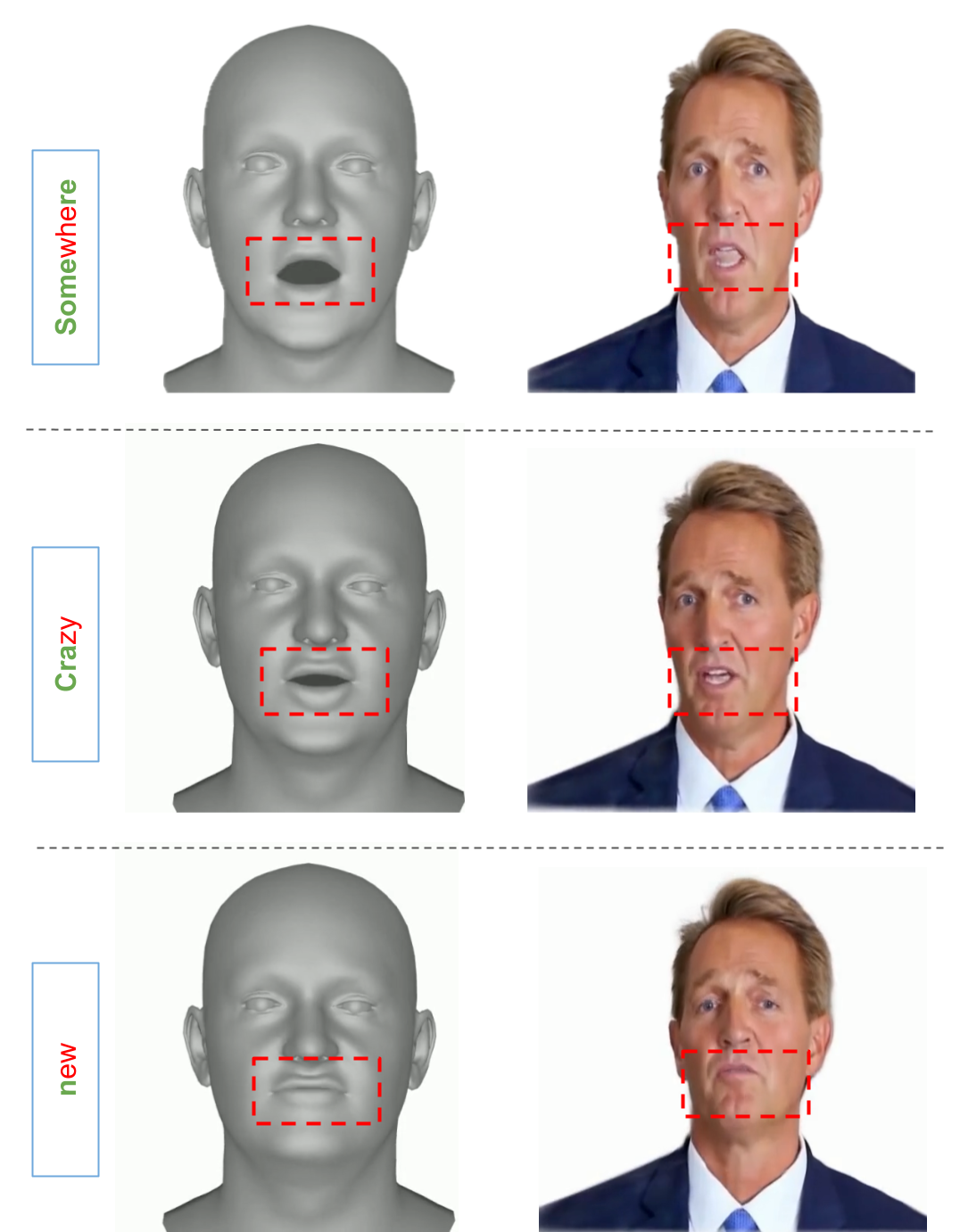}
  \caption{For a given audio signal, GaussianHeadTalk generates a lip-sync 3D mesh and use the generated FLAME parameters to transfer lip motion on a trained GaussianAvatar with optimized FLAME parameters.}
  \label{fig:motion_transfer_from_mesh}
\end{figure}

\section{Qualitative Ablation Study}
\begin{figure}[h!]
  \centering
  \includegraphics[width=\linewidth]{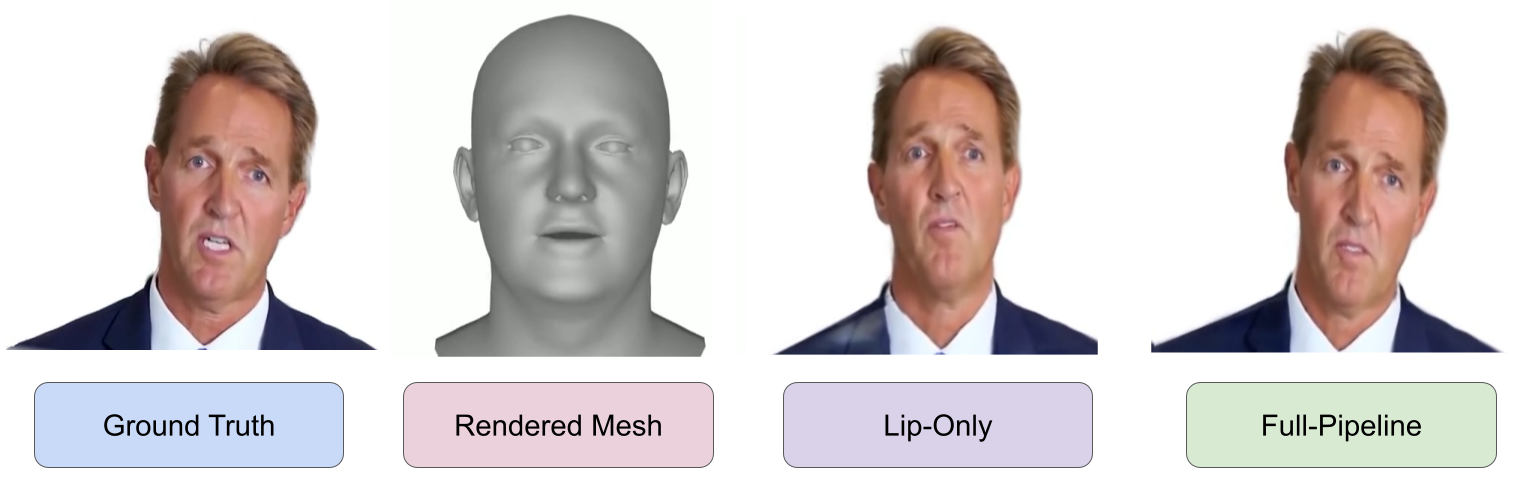}
  \caption{Ablation Study: Effect of transferring the lip motion and keeping other parameters static (w/o Full Motion Transfer). The results shows visible artifacts in the generated avatar, as the FLAME parameters are not fully independent.}
  \label{fig:Ablation_HeadMotion}
\end{figure}

\begin{figure}[h!]
  \centering
  \includegraphics[width=\linewidth]{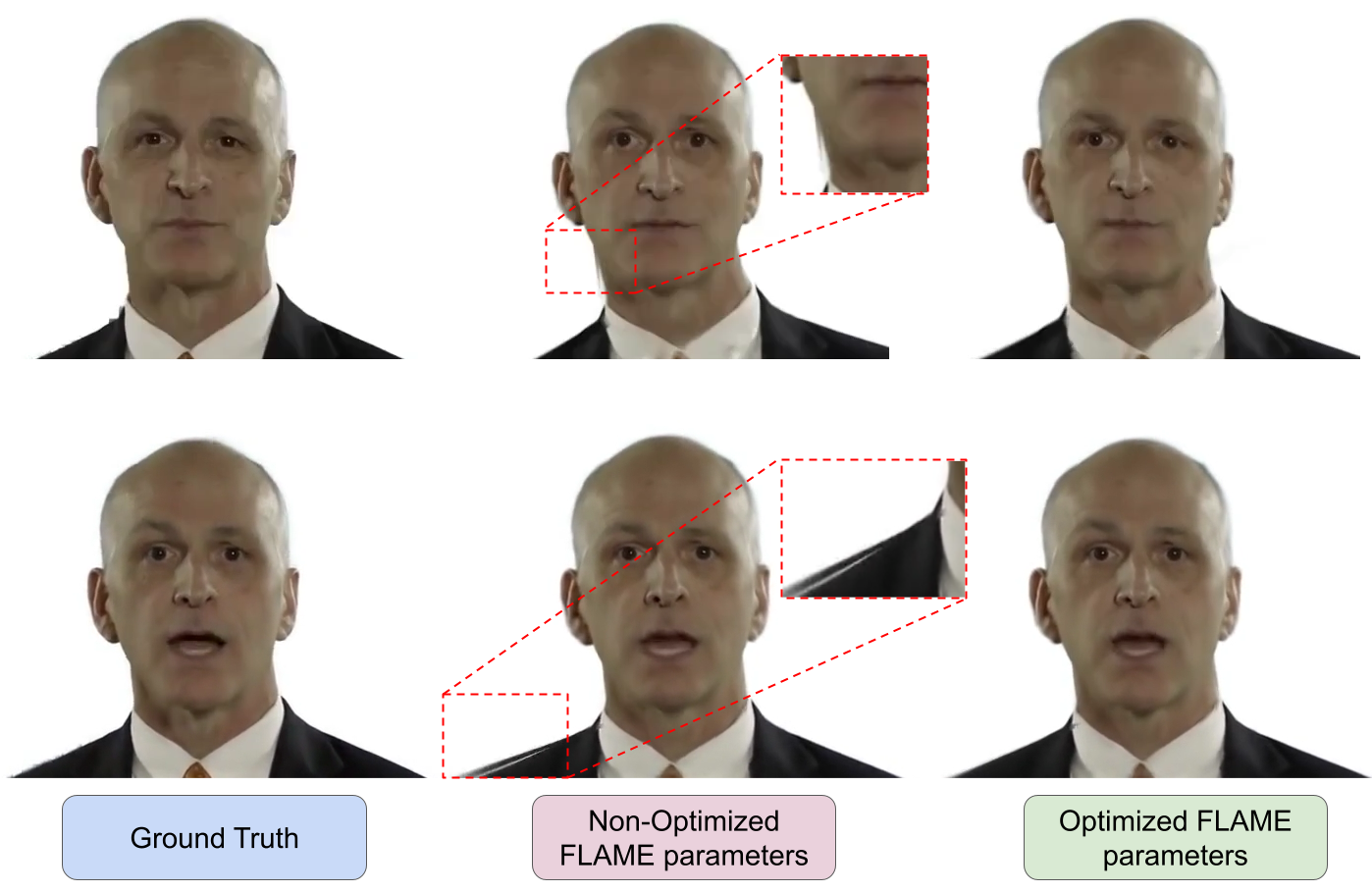}
  \caption{Ablation Study: Using Non-Optimized FLAME parameters (w/o Parameter Optimization). This leads to artifacts around the torso region, and wobbling issues. }
  \label{fig:ablation-keypoints-optimization}
\end{figure}

\section{Temporal Analysis of Keypoints}
\begin{figure}[hb]
  \centering
  \includegraphics[width=\linewidth,height=150pt]{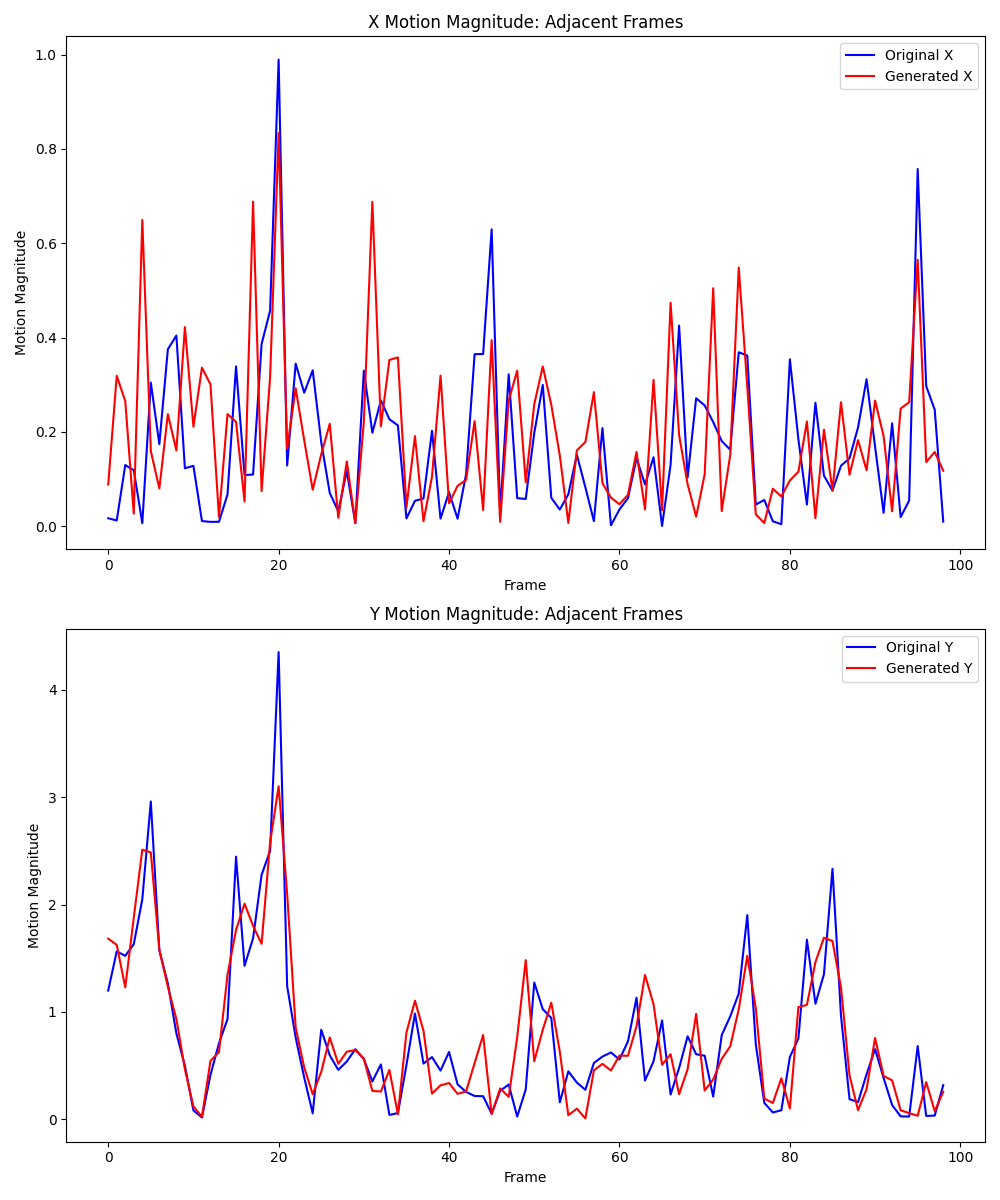}
  \caption{Comparison of keypoint movement across time between an original video and a video generated using GaussianTalker. The overlay graph shows that there is flickering in the rendered video. In ideal case, these two graphs should be perfectly overlapping. We report $x$-axis, $y$-axis motion magnitude over time in upper and lower plots, respectively.}
\end{figure}

\section{User Study}
\begin{table*}[h!]
\centering
\begin{tabular}{@{}llccccc@{}}
\toprule
\textbf{Category} & \textbf{Method} & \textbf{Final Score} & \textbf{`Best' (3)} & \textbf{`Average' (2)} & \textbf{`Worst' (1) } & \textbf{Total Ratings} \\ \midrule
\multirow{3}{*}{\textbf{Naturalness}} & \textbf{GaussianHeadTalk (ours)} & \textbf{9.8} & 284 & 14 & 2 & 300 \\
 & TalkingGaussian~\cite{li2024talkinggaussian} & 6.2 & 40 & 178 & 82 & 300 \\
 & GaussianTalker~\cite{cho2024gaussiantalker} & 4.0 & 5 & 50 & 245 & 300 \\ \midrule
\multirow{3}{*}{\textbf{LipSync}} & \textbf{GaussianHeadTalk (ours)} & \textbf{7.8} & 142 & 118 & 40 & 300 \\
 & TalkingGaussian~\cite{li2024talkinggaussian} & 7.2 & 128 & 92 & 80 & 300 \\
 & GaussianTalker~\cite{cho2024gaussiantalker} & 5.0 & 25 & 100 & 175 & 300 \\ \midrule
\multirow{3}{*}{\textbf{Quality}} & \textbf{GaussianHeadTalk (ours)} & \textbf{9.5} & 260 & 35 & 5 & 300 \\
 & TalkingGaussian~\cite{li2024talkinggaussian} & 6.5 & 80 & 125 & 95 & 300 \\
 & GaussianTalker~\cite{cho2024gaussiantalker} & 4.0 & 1 & 58 & 241 & 300 \\ \bottomrule
\end{tabular}
\label{tab:requested_scores_breakdown}
\caption{Detailed breakdown of User Study Ratings. 30 participants evaluate 10 videos of each method, generating 300 ratings in total. For each triplet, a participant assigns the ranks 1, 2, and 3 once each, so the raw sum across the three methods is 1+2+3=6. Over 10 triplets, this gives a total of 10×6=60. After dividing each method’s total by 3 for normalization, the overall sum across all methods is fixed at 60/3 = 20.}
\end{table*}

\begin{figure*}[t!]
  \centering
  \includegraphics[width=\linewidth]{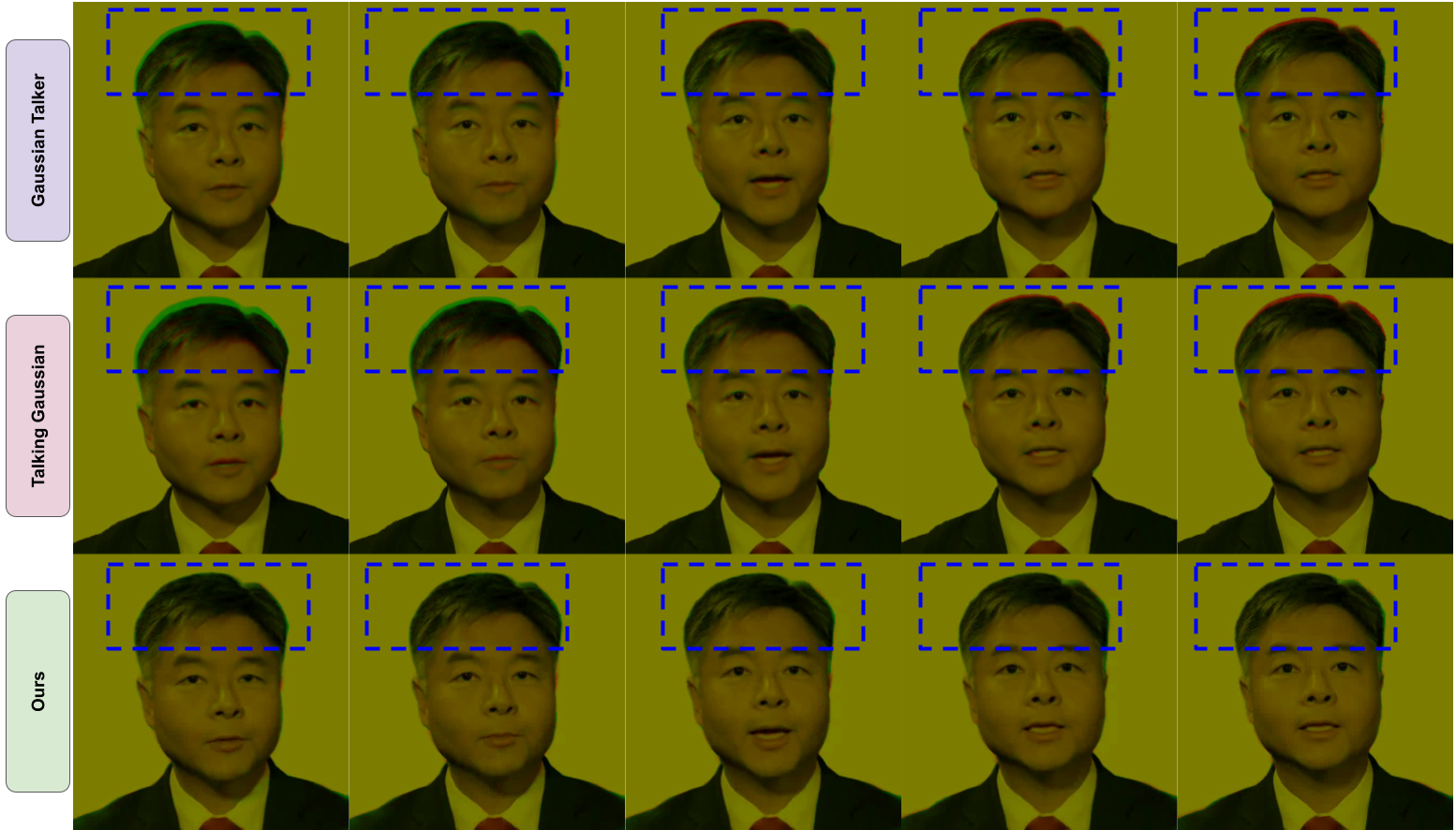}
  \caption{Visualization of frame stability through color channel overlay with ground-truth video over 10 consecutive frames. The significant displacement (wobbling) observed in the GaussianTalker and TalkingGaussian methods contrasts with the high overlap and stability achieved by our proposed method. Green and red channels highlight the differences within the blue dashed boxes.}

  \label{fig:wobbling-overlay-supp}
\end{figure*}

\begin{figure*}[htbp]
  \centering
  \includegraphics[width=\linewidth,height=550pt]{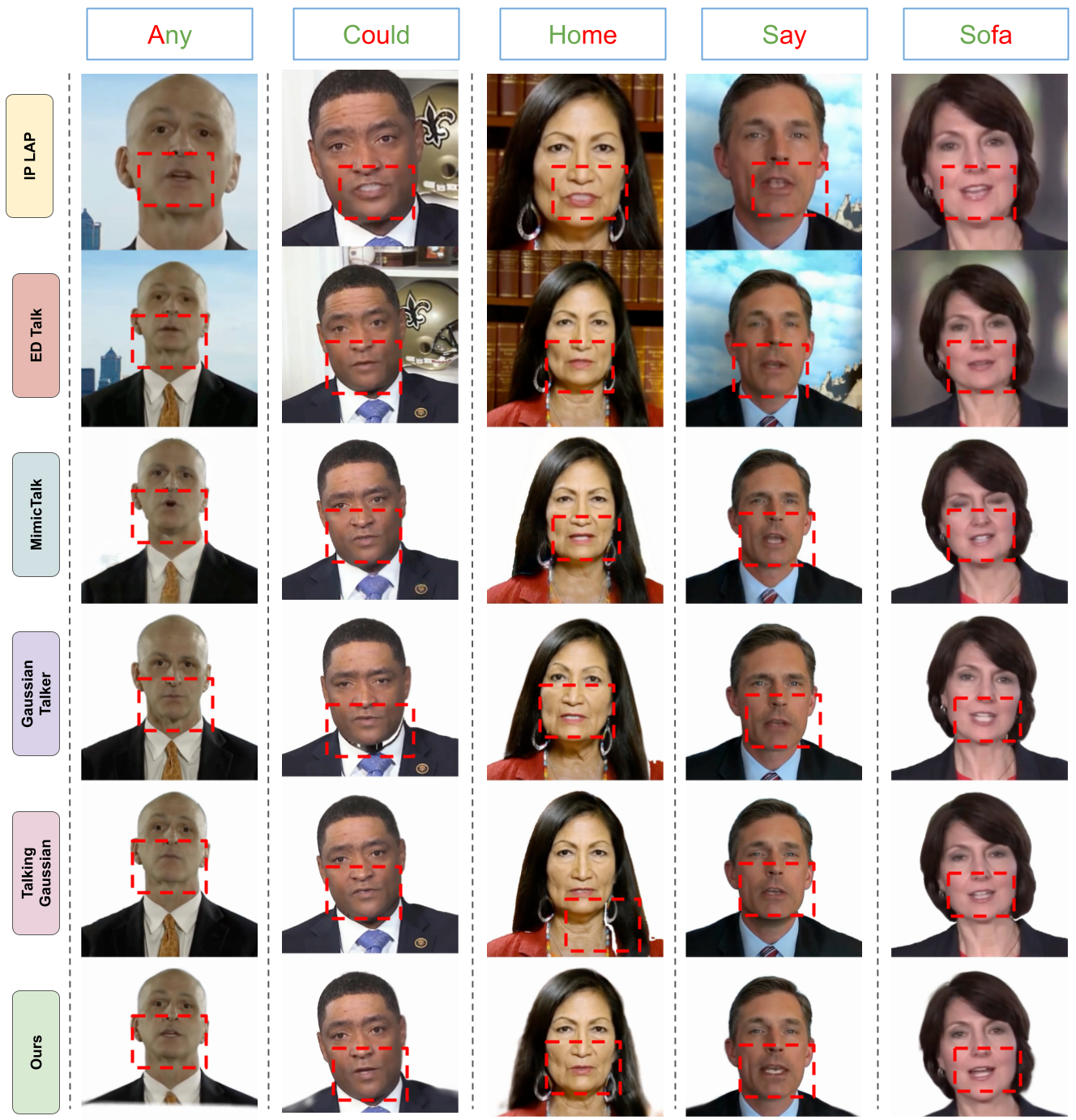}
  \caption{Cross-Reenactment Results: We show the visual results by reenacting various methods using a different audio, from a different speaker. The top row shows the word from the audio, with red part highlighting the exact phoneme. GaussianHeadTalk provides the best possible lip movement for these new audio samples. Other methods struggle to have proper lip motion, generate high-quality videos and no artifacts.}

  \label{fig:qualitative-overlay-supp-1}
\end{figure*}

\begin{figure*}[htbp]
  \centering
  \includegraphics[width=\linewidth,height=550pt]{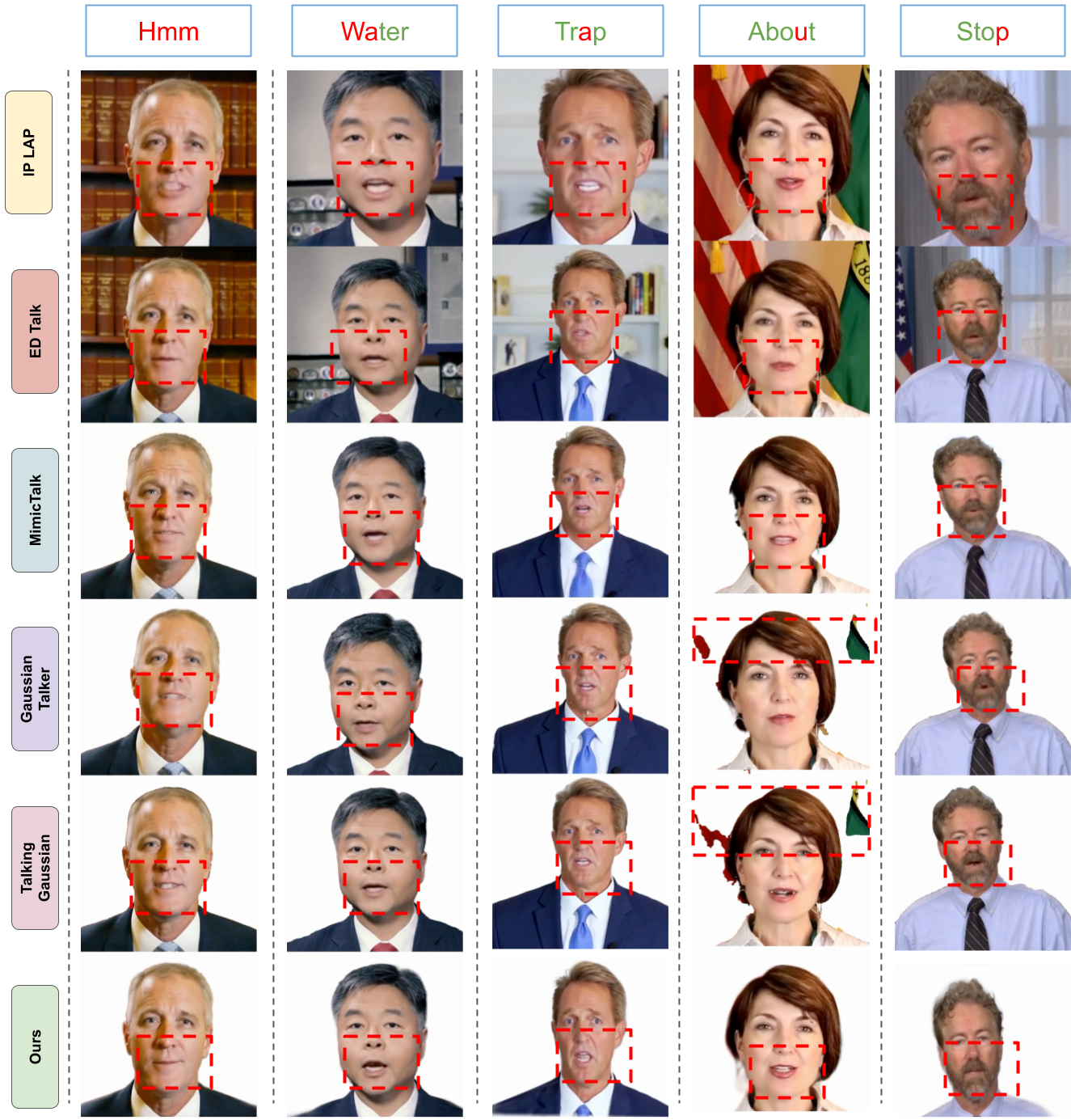}
  \caption{Cross-Reenactment Results: We show the visual results by reenacting various methods using a different audio, from a different speaker. The top row shows the word from the audio, with red part highlighting the exact phoneme. GaussianHeadTalk provides the best possible lip movement for these new audio samples. Other methods struggle to have proper lip motion, generate high-quality videos and no artifacts.}

  \label{fig:qualitative-overlay-supp-2}
\end{figure*}

\end{document}